\documentclass[sigconf]{acmart}
\setcopyright{none}
\acmConference[]{}{}{}
\acmBooktitle{} 
\acmISBN{} 
\acmDOI{}
\acmYear{} 
\acmPrice{}
\renewcommand{\footnote}[1]{}
\acmVolume{} 
\acmNumber{} 
\acmArticle{} 
\usepackage{placeins}
\usepackage{array}
\usepackage{graphicx}
\AtBeginDocument{%
  }

\begin{document}

\title{Decoding News Bias: Multi Bias Detection in News Articles}

\author{Bhushan Santosh Shah}
\email{shahbs21.comp@coeptech.ac.in}
\affiliation{%
  \institution{College of Engineering, Pune Technological University}
  \country{India}
}

\author{Deven Santosh Shah}
\email{dsshah@cs.stonybrook.edu}
\affiliation{%
  \institution{Stony Brook University}
  \country{USA}
}

\author{Vahida Attar}
\email{vahida.comp@coeptech.ac.in}
\affiliation{%
  \institution{College of Engineering, Pune Technological University}
  \country{India}
}

\begin{abstract}
News Articles provides crucial information about various events happening in the society but they unfortunately come with different kind of biases. These biases can significantly distort public opinion and trust in the media, making it essential to develop techniques to detect and address them. Previous works have majorly worked towards identifying biases in particular domains e.g., Political, gender biases. However, more comprehensive studies are needed to detect biases across diverse domains. Large language models (LLMs) offer a powerful way to analyze and understand natural language, making them ideal for constructing datasets and detecting these biases. In this work, we have explored various biases present in the news articles, built a dataset using LLMs and present results obtained using multiple detection techniques. Our approach highlights the importance of broad-spectrum bias detection and offers new insights for improving the integrity of news articles. 
\end{abstract}




\maketitle

\section{Introduction}
Over the last decade, rapid technological advancements and increased internet accessibility have led to a significant shift in how information is distributed, with a growing preference for online news articles over traditional print media \citep{krieger2022domain, gonccalves2021local}. Today, news articles serve as a vital source of information for millions of people globally \cite{hamborg2019automated}.
They play a crucial role in spreading awareness, educating the public, and providing real-time updates on various events. As such, they significantly influence public opinion, shaping how people perceive societal, political, and global matters \cite{hamborg2019automated}. This power to inform and influence makes the integrity and impartiality of news articles critically important. 

However, news articles are often subject to various forms of media bias. These biases can skew the information presented, leading to misleading narratives and polarizing views. Biases in news articles may manifest in several forms, including \textbf{political bias}, where certain political parties or ideologies are unreasonably favored; \textbf{gender bias}, which portrays or criticizes individuals based on gender-related attributes; \textbf{entity bias}, favoring specific organizations or people; \textbf{racial bias}, where content favors or marginalizes certain racial or cultural groups and other kinds of biases. The presence of such biases can significantly distort the objectivity of news, misguiding readers and perpetuating misinformation.

Detecting biases in news articles is essential for promoting transparency and fostering critical thinking. Identifying these biases allows readers to better evaluate the information they consume, making informed decisions rather than being influenced by skewed content. Moreover, news article providers can benefit by improving their content's credibility and gaining the trust of a wider audience. By highlighting biases, they can maintain higher ethical standards and work towards providing more balanced and objective reporting.

In recent years, Large Language Models (LLMs) have demonstrated impressive capabilities in understanding and generating natural language \cite{naveed2023comprehensive}. These advancements have opened new possibilities in analyzing and processing text at an unprecedented scale and accuracy. However, leveraging LLMs specifically for detecting biases in news reporting remains relatively underexplored. 

This study make following key contributions:
\begin{itemize}
    \item \textbf{Extending the Scope:} We broaden the scope of the bias detection problem by identifying and analyzing different types of biases that may occur in news articles.
    \item \textbf{Dataset Annotation Method:} We introduce a preliminary method for dataset annotation using LLMs, recognizing that further research is required to thoroughly assess the reliability and consistency of LLM-driven annotations.
    \item \textbf{Experimental Results:} We experimented with different transformer based models on the curated dataset to assess their effectiveness in identifying news biases, providing a comparative analysis of existing methods.
\end{itemize}

The rest of this paper is structured as follows: Section 2 presents a detailed review of the related works in bias detection and the use of LLMs for natural language tasks. Section 3 discusses our methodology for bias detection, including how we build and annotate the dataset. Section 4 details the experiments conducted and the corresponding results derived from the techniques applied to the dataset. Section 5 offers the concluding remarks of the study and lastly, we address the study's limitations and explores potential future research directions.

\section{Related Works}
Recent studies have predominantly focused on detecting political bias and media bias in news articles. Media bias, particularly bias by word choice and framing, has been a central topic of research. For instance, \citeauthor{krieger2022domain} introduced DA-RoBERTa, a domain-adaptive model that detects media bias, particularly bias by word choice, achieving state-of-the-art performance in detecting biased language through a domain-adaptive pre-training approach. Similarly, Gaussian Mixture Models (GMM) have been employed to assess article-level bias by analyzing sentence-level features, such as bias frequency and bias sequence, effectively capturing the probabilistic nature of bias \cite{chen2020detecting}.

Further research has explored linguistic and context-oriented features to identify subtle word-level bias in media leveraging demographic insights of annotators to enhance bias detection \cite{SPINDE2021102505}. Additionally, \citeauthor{AGGARWAL2020100025} have gone beyond mere detection and proposed systems to assess the short-term impact of media bias on public opinion, as seen in research analyzing biased reporting in the context of Twitter posts by Indian media outlets.

Political bias detection research has largely aimed at uncovering partisan leanings in news content. For example, a framework has been used to analyze the framing and structure of news across sources, showing improved performance over traditional models \cite{nadeem2021detecting}. Another method incorporated headline attention mechanisms to detect bias, enhancing accuracy with neural networks \cite{gangula2019detecting}. Additionally, adversarial media adaptation has focused on emphasizing content over source bias, improving the prediction of political ideologies in news articles \cite{baly2020we}.

In recent years, there has been a growing interest in incorporating Large Language Models (LLMs) into bias detection systems to enhance their capabilities. One prominent example is BiasScanner, an application that leverages a pre-trained LLM to detect biased sentences in news articles and provides explanations for its decisions \citep{menzner2024biasscanner}.

One study explored GPT-4's ability to classify political bias from web domains, showing strong correlation with human-verified sources like Media Bias/Fact Check (MBFC). While promising for scalable bias detection, GPT-4 abstained from classifying less popular sources, highlighting the need for combining LLMs with human oversight \cite{hernandes2024llms}. Another study used GPT-3.5 for annotating politically biased news articles, blending LLMs with expert-driven rules for identifying bias indicators \cite{raza2024dataset}.

Moreover, there are a few studies which have incorporated LLMs in detecting fake news \cite{boissonneault2024fake, teo2024integrating, jiang2024disinformation}. While fake news detection and bias detection seem similar, they address distinct issues. Fake news detection focuses on identifying completely false or misleading content designed to deceive, using fact-checking or external validation. In contrast, bias detection uncovers partiality or slant in how factual information is framed.

However, while these effective in specific bias detection tasks, there  is still a gap in exploring diverse types of biases. Most studies focus on singular forms of bias, such as political or word-choice bias, without addressing a more comprehensive range of biases that may exist in news articles. Additionally, none of these approaches have leveraged LLMs to build a multi-label dataset capable of detecting various types of biases in a single framework. Given the limitations of prior work, our research proposes a new approach that explores multiple forms of bias using LLMs, thus enabling a broader, multi-label bias detection framework that has yet to be fully explored in the literature.

\section{Methodology}
In this section, we will first outline the various types of biases along with their definitions to ensure a clear understanding of each bias. Afterward, we will delve into the process of dataset creation, discuss how the labels were extracted, and detail the techniques applied to the collected and filtered data.
\subsection{Bias Definitions}
In prior studies, labeling datasets often relied on simplified techniques, such as left-center-right categorization or a binary biased-unbiased approach. However, these methods overlook the more nuanced and distinct types of biases present in news content. To address this, we chose to label each major form of bias individually, ensuring a more detailed and accurate representation. Below is a list of the various biases along with their definitions:
\begin{figure*}[t]
  \centering
  \includegraphics[width=\textwidth]{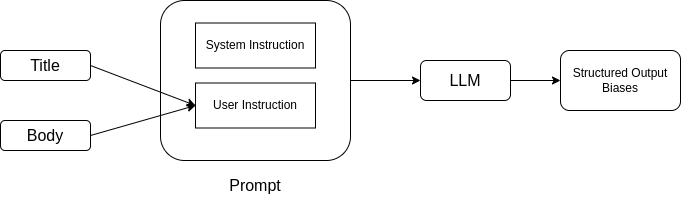} 
  \caption{Label Extraction Flow.}
  \label{fig:label_extraction}
\end{figure*}

\begin{figure*}[t]
  \centering
  \begin{minipage}{0.32\textwidth}
    \includegraphics[width=\linewidth]{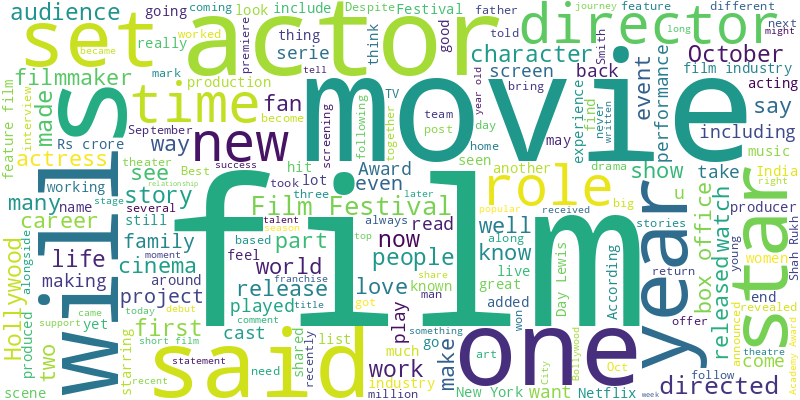}
    \caption{Hollywood Wordcloud}
    \label{fig:hollywood_wordcloud}
  \end{minipage}
  \hfill
  \begin{minipage}{0.32\textwidth}
    \includegraphics[width=\linewidth]{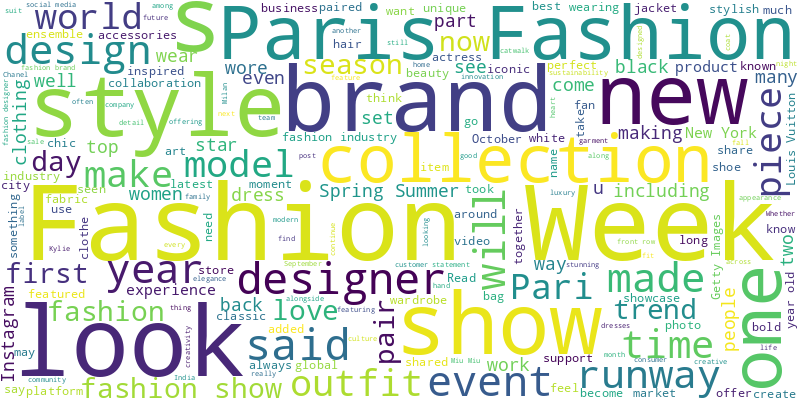}
    \caption{Fashion Wordcloud}
    \label{fig:fashion_wordcloud}
  \end{minipage}
  \hfill
  \begin{minipage}{0.32\textwidth}
    \includegraphics[width=\linewidth]{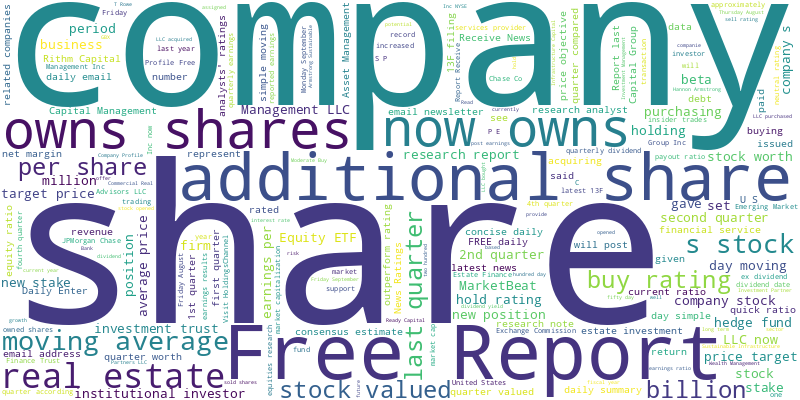}
    \caption{Finance Wordcloud}
    \label{fig:finance_wordcloud}
  \end{minipage}
  
  \vspace{1em} 

  \begin{minipage}{0.32\textwidth}
    \includegraphics[width=\linewidth]{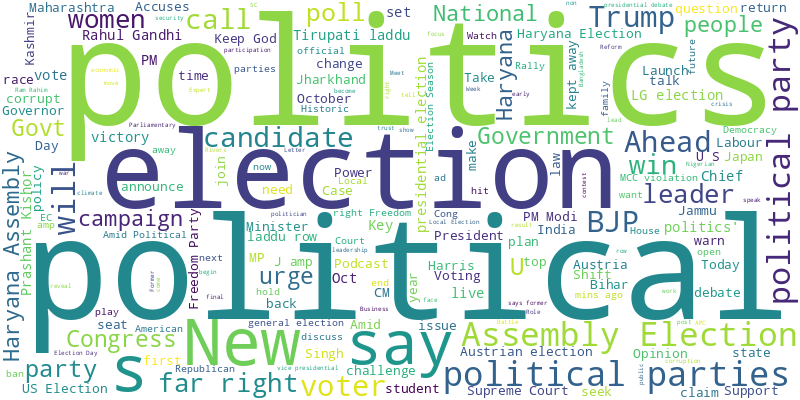}
    \caption{Politics Wordcloud}
    \label{fig:politics_wordcloud}
  \end{minipage}
  \hfill
  \begin{minipage}{0.32\textwidth}
    \includegraphics[width=\linewidth]{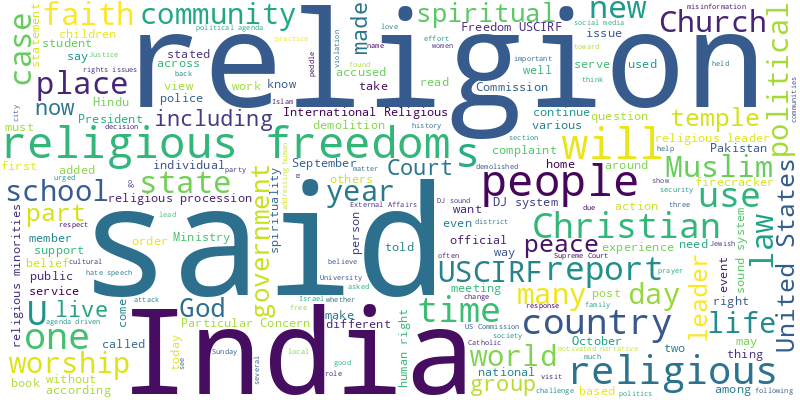}
    \caption{Religion Wordcloud}
    \label{fig:religion_wordcloud}
  \end{minipage}
  \hfill
  \begin{minipage}{0.32\textwidth}
    \includegraphics[width=\linewidth]{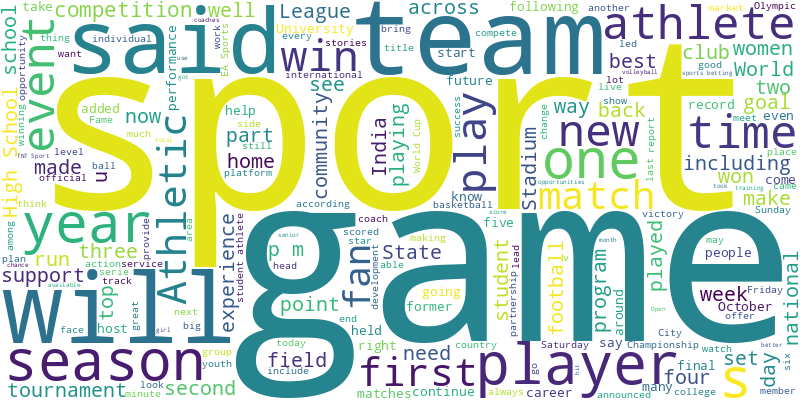}
    \caption{Sports Wordcloud}
    \label{fig:sports_wordcloud}
  \end{minipage}
\end{figure*}

\begin{itemize}
    \item \textbf{Political Bias:} This refers to articles that unreasonably favor or criticize a political party, ideology, or government policy.
    \item \textbf{Gender Bias:} This bias occurs when individuals or groups are evaluated or treated based on their gender, particularly with a focus on appearances or stereotypical roles.
    \item \textbf{Entity Bias:} Entity bias manifests when reporting disproportionately criticizes or praises specific individuals, corporations, or other entities, regardless of objective facts.
    \item \textbf{Racial Bias:} This is evident when articles favor or disfavor individuals or groups based on race, nationality, ethnicity, or culture.
    \item \textbf{Religious Bias:} The unfair favoring or critique of a particular religion or its followers. News articles may show religious bias by unfairly portraying certain faiths as superior or inferior, or by emphasizing the actions of specific religious groups in a misleading way.
    \item \textbf{Regional Bias:} This bias occurs when individuals or events are depicted unfairly based on their geographic location, often leading to unequal or skewed coverage.
    \item \textbf{Sensationalism:} The use of exaggerated or shocking headlines and content to attract attention, often at the expense of factual accuracy. This bias prioritizes emotional appeal and drama over balanced, objective reporting, potentially distorting the reader’s perception of the event.
\end{itemize}

\begin{figure*}[t] 
    \centering
    \begin{minipage}{0.48\textwidth} 
        \centering
        \includegraphics[width=\linewidth]{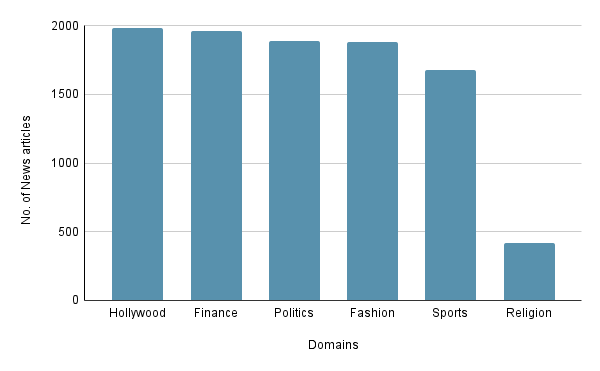}
        \caption{News Articles Count per Domain}
        \label{fig:domain_count}
    \end{minipage}
    \hfill
    \begin{minipage}{0.48\textwidth} 
        \centering
        \includegraphics[width=\linewidth]{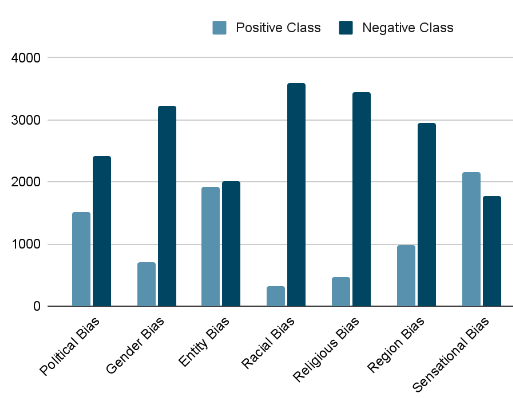}
        \caption{Count of Positive and Negative Classes for each Bias}
        \label{fig:classes_count}
    \end{minipage}
\end{figure*}

\subsection{Dataset Collection}
To develop a robust model for bias detection, we curated a diverse dataset of news articles across six key domains: \textbf{Hollywood}, \textbf{Fashion}, \textbf{Finance}, \textbf{Religion}, \textbf{Politics} and \textbf{Sports}.

These domains were deliberately selected because we believe they are more likely to exhibit one or more types of biases, making them ideal for testing the model’s ability to detect biases across different contexts. This diversity ensures that the model is exposed to varied instances of bias, enhancing its generalizability and performance. We utilized an API provided by Aylein (https://aylien.com/) to gather full-text news articles. Figure~\ref{fig:domain_count} shows the number of news articles fetched for each of the domains. A total of 9790 articles were collected, with approximately 1850 articles from each domain except religion domain which has only 415 articles. We believe this is because religion is a highly niche and personal subject and therefore not a lot of news articles are published as compared to other domains. We labeled the dataset as discussed in Section~\ref{sec:label_extraction}. We filtered the dataset by only keeping those examples which had at least one bias from 7 bias categories and discarded others. This process reduced the dataset size to 4886 samples.

Additionally, Figures~\ref{fig:hollywood_wordcloud} to \ref{fig:sports_wordcloud} presents word clouds for each of the domains, offering a visual summary of the common themes and frequently used words in the news articles. These word clouds provide a quick overview of the linguistic patterns and key topics within each domain, helping us better understand the nature of the articles.
\subsection{Label Extraction}
\label{sec:label_extraction}
Manually labeling a large dataset for various biases is both labor-intensive and prone to inconsistencies. To address this challenge and automate the labeling process, we propose leveraging a Large Language Model (LLM) with increased language understanding capabilities \cite{naveed2023comprehensive}. The overall flow of the label extraction process is illustrated in Figure~\ref{fig:label_extraction}.

\textbf{LLM Selection:} We selected GPT-4o mini for its strong performance in language understanding tasks, alongside its cost-effectiveness, making it suitable for labeling large datasets. Its capacity to accurately interpret nuanced language constructs ensures both consistency and reliability in the generated outputs.

\textbf{Prompting Techniques:} Several prompting techniques exist for LLMs, including zero-shot prompting, few-shot prompting \cite{brown2020language}, instruction-based prompting, chain-of-thought prompting \cite{wei2022chain}, and contextualized prompting with external knowledge \cite{lewis2020retrieval}. In this study,  we opted for instruction-based prompting due to its simplicity and feasibility, especially given the cost constraints associated with the total number of prompt tokens. The prompt was designed with clear and structured user and system instructions to ensure reliable results.

\textbf{User and System Instruction:} We provided the model with a detailed system instruction that included specific definitions for each type of bias. The instruction was designed to clearly outline the task and how the results should be formatted. The bias definitions were included in the instruction, and an example output format was provided to guide the model. Additionally, the title and body of each news article were passed through the user instruction. The prompt template and its description is given in the Appendix \ref{appendix:a}. 
Using this approach, the GPT-4omini model labeled the entire dataset of 9790 articles, drawn from six distinct domains. The entire process took approximately 6 hours, providing an efficient, scalable, and consistent labeling mechanism across the dataset.

\begin{table*}[htbp!]
\small
\centering
\resizebox{\textwidth}{!}{%
\begin{tabular}{l|ccc|ccc|ccc|ccc|ccc}
\hline
\textbf{Bias} & \multicolumn{3}{c|}{\textbf{BERT}} & \multicolumn{3}{c|}{\textbf{DistilBERT}} & \multicolumn{3}{c|}{\textbf{ALBERT}} & \multicolumn{3}{c|}{\textbf{RoBERTa}} & \multicolumn{3}{c}{\textbf{XLNet}} \\ \hline
 & \textbf{P} & \textbf{R} & \textbf{F1} & \textbf{P} & \textbf{R} & \textbf{F1} & \textbf{P} & \textbf{R} & \textbf{F1} & \textbf{P} & \textbf{R} & \textbf{F1} & \textbf{P} & \textbf{R} & \textbf{F1} \\ \hline
Political Bias   & \textbf{0.86} & 0.93 & \textbf{0.89} & 0.82 & 0.93 & 0.87 & 0.79 & 0.90 & 0.84 & 0.82 & \textbf{0.94} & 0.87 & 0.80 & 0.89 & 0.84 \\ 
Gender Bias      & \textbf{0.73} & 0.78 & \textbf{0.75} & 0.66 & 0.73 & 0.70 & 0.58 & 0.75 & 0.66 & 0.52 & \textbf{0.81} & 0.63 & 0.52 & \textbf{0.81} & 0.64 \\ 
Entity Bias      & \textbf{0.75} & 0.74 & 0.74 & 0.71 & 0.77 & 0.74 & 0.73 & 0.73 & 0.73 & 0.72 & \textbf{0.79} & \textbf{0.75} & 0.71 & 0.78 & \textbf{0.75} \\ 
Racial Bias      & \textbf{0.60} & 0.64 & \textbf{0.62} & 0.48 & 0.65 & 0.55 & 0.28 & 0.71 & 0.40 & 0.25 & 0.71 & 0.37 & 0.26 & \textbf{0.72} & 0.38 \\ 
Religious Bias   & \textbf{0.83} & \textbf{0.96} & \textbf{0.89} & 0.81 & 0.92 & 0.86 & 0.62 & 0.94 & 0.74 & 0.65 & 0.93 & 0.76 & 0.66 & 0.95 & 0.78 \\ 
Region Bias      & \textbf{0.61} & 0.75 & \textbf{0.67} & 0.49 & 0.65 & 0.56 & 0.46 & 0.59 & 0.52 & 0.40 & \textbf{0.77} & 0.53 & 0.39 & 0.73 & 0.51 \\ 
Sensational Bias & 0.74 & \textbf{0.86} & \textbf{0.80} & \textbf{0.78} & 0.65 & 0.71 & 0.72 & 0.78 & 0.75 & 0.76 & 0.67 & 0.71 & 0.74 & 0.67 & 0.70 \\ 
\end{tabular}%
}
\caption{Performance comparison of various models (BERT, DistilBERT, ALBERT, RoBERTa, and XLNet) across different bias types using Precision, Recall, and F1 Score metrics.}

\label{tab:results}

\end{table*}

\section{Experiments}
In this study, we evaluated several transformer-based models to classify biases in news articles, specifically focusing on BERT \cite{devlin2018bert}, RoBERTa \cite{liu2019roberta}, ALBERT \cite{lan2019albert}, DistilBERT \cite{sanh2019distilbert}, and XLNet \cite{yang2019xlnet}. One major challenge was the significant class imbalance present in the dataset, as illustrated in Figure~\ref{fig:classes_count}, where the number of samples for the positive class is substantially lower than that for the negative class across most biases. To address this issue, we implemented an inverse frequency weighting method, which assigned higher weights to the positive class for each label during training, helping to mitigate the effects of class imbalance.

Experiments were conducted on a T4 GPU using a batch size of 8, leveraging the AdamW optimizer \cite{loshchilov2017decoupled}, a commonly used variant of the Adam optimizer, designed to better handle overfitting through weight decay. The linear learning rate scheduler was configured with a starting learning rate of  \(2 \times 10^{-5}\), which decays over time to final learning rate to 0, ensuring smooth convergence. Each model was trained for 6 epochs, balancing between model performance and training time efficiency.

In multilabel classification, it is particularly important to retain the distribution of all labels across folds, as each sample can be associated with multiple labels simultaneously. A crucial aspect of this study was the dataset splitting strategy. Rather than randomly splitting the dataset into training, validation, and test sets, which could lead to skewed splits and unrepresentative distributions of the various labels, we used Multilabel Stratified KFold splitting. This method ensures that the label distribution is proportionally maintained across the training, validation, and test sets, preventing the model from learning from biased splits.
\section{Results}
Following our experiments with various transformer-based models—BERT, RoBERTa, ALBERT, DistilBERT, and XLNet—we evaluated each model’s capacity to classify different types of biases present in news articles. Table~\ref{tab:results} provides a summary of the models' performance across the biases, measured through precision, recall, and F1-score. Our analysis highlights both the models' effectiveness in detecting explicit bias types and the challenges posed by class imbalance and nuanced biases.
\begin{itemize}
    \item \textbf{Model Performance Comparison} - BERT consistently outperformed other models across most bias categories, particularly achieving an F1-score of 0.89 in Political Bias detection, which may be attributed to BERT's robust contextual embedding capabilities. RoBERTa showed a competitive performance, closely following BERT in several categories. However, models like DistilBERT and ALBERT, with fewer parameters, generally exhibited lower F1-scores, particularly in Gender and Sensational Bias detection. 
    \item \textbf{Class Imbalance Impact and Mitigation} - Class imbalance posed a considerable challenge in our dataset, especially in Racial and Region Bias detection, which had notably fewer positive samples. Despite our use of inverse frequency weighting to emphasize the minority classes, models like DistilBERT and XLNet struggled with these biases, achieving F1-scores as low as 0.38 and 0.51 in Racial Bias detection, respectively. This imbalance suggests that additional data or more sophisticated augmentation techniques may be necessary to further alleviate these performance gaps.
    \item \textbf{LLM Annotations Reliability} - While GPT-4omini provided fairly accurate annotations, we identified several errors in its assessments. In some cases, it incorrectly flagged articles as biased when they did not contain bias particularly in articles with multiple biases. Appendix \ref{appendix:b} includes examples of these misannotations. Improving the accuracy of LLM annotations remains an area of future exploration.
\end{itemize}

\section{Conclusion}
In this study, we expanded the framework for bias detection in news articles by identifying various types of biases rather than simply categorizing them as biased or unbiased. Our unique dataset, labeled using large language models (LLMs), represents the first attempt to classify news articles according to seven distinct bias categories prevalent in the media landscape. We applied transformer-based models to this dataset and conducted a comparative analysis of their performance, concluding that BERT emerged as the most effective model overall. This research provides valuable insights into the complexities of bias detection in news media, paving the way for more nuanced approaches in future studies.

\section*{Limitations and Future Work}
We utilized an LLM, specifically GPT-4omini, to label the dataset, which significantly reduced the time required for annotation. However, relying solely on the LLM for labeling can introduce some discrepancies and inconsistencies. In future work, we aim to explore methods to balance annotation reliability while maximizing the LLM's capabilities.

Another limitation lies in the dataset itself, particularly the imbalance of samples across bias categories. Although techniques were employed to mitigate the impact of class imbalance, exploring advanced methods to expand and diversify the dataset—especially for underrepresented biases—could lead to more accurate model performance.

Additionally, the dataset splitting method could be further refined. While Multilabel Stratified KFold was used to maintain label distribution, future work could investigate other splitting strategies to prevent potential skewness and better reflect real-world distributions. This, combined with further experimentation on alternative data augmentation techniques, could improve the generalization capabilities of the models.
Addressing these limitations would strengthen the reliability of the system and contribute to more accurate and comprehensive bias detection in news articles.

\appendix

\FloatBarrier 
\begin{figure*}[t!] 
    \centering
    \begin{minipage}{\textwidth} 
        \centering
        \includegraphics[width=\linewidth]{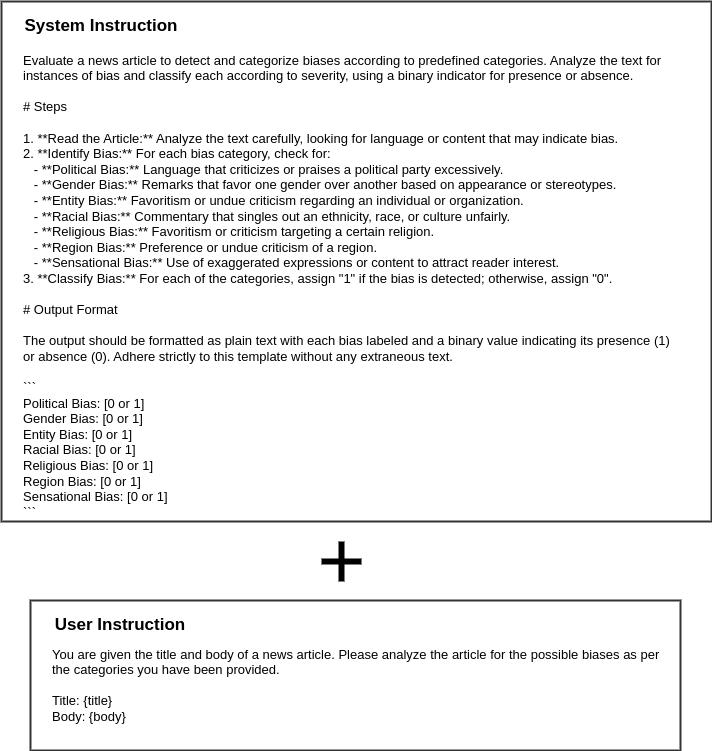}
        \caption{LLM prompt}
        \label{fig:prompt}
    \end{minipage}
\end{figure*}
\section{Prompt}
\label{appendix:a}
\textbf{Prompt}: The prompt to the LLM consisted of system and user instructions for evaluating news articles for potential biases based on predefined categories. System instructions outline a structured approach for analyzing the text to identify various bias types such as political, gender, entity, racial, religious, regional, sensational, rumor, and inconsistent content. Each bias is classified by severity, using a binary indicator (1 for presence, 0 for absence), ensuring systematic and consistent bias detection. User instruction provide the title and body of a news article as input to the LLM, and the model applies the outlined prompt to detect and categorize biases according to the specified criteria, with the output presented in a clearly defined format that labels each bias category with its corresponding binary value, in strict adherence to the provided output format.

\section{LLM Annotations Analysis}
\label{appendix:b}
In order to check the reliability of the annotations given by the LLM, we performed a qualitative analysis by reviewing a set of articles and their corresponding annotations. Overall, we found that the LLM did a good job assigning most of the biases correctly, especially in cases of gender, entity, racial, and sensational biases. These types of biases are generally more straightforward to identify due to the inherent language used in the articles, making them easier for the LLM to detect.

However, we identified some inconsistencies in the classification of religious, regional, and political biases. Specifically, certain articles that mentioned religious practices, regional dynamics, or political figures were misclassified as exhibiting bias, even though the content itself did not demonstrate overt prejudice or discriminatory intent. Examples of the LLM misannotions are provided in Table ~\ref{tab:annotations}.

For example, articles discussing religious events or political meetings were flagged as having religious or political bias, even though the content merely provided factual reporting without taking a stance or making value judgments. Similarly, articles about regional conflicts or sports conference realignments were misclassified as showing regional bias, despite the focus being on logistical or structural topics, rather than promoting or disparaging specific regions.

These misclassifications likely result from the LLM’s sensitivity to specific keywords (e.g., "religious ritual" or "political figures") that may trigger assumptions of bias, even in neutral contexts. Similarly, discussions of historical tensions or regional fit could be misinterpreted as regional bias, despite their focus on practical matters rather than promoting divisions.

Overall, while the LLM performs well in identifying more overt forms of bias, the occasional misclassification suggests the need for a more nuanced understanding of context, especially when dealing with topics where certain terms or events may be misinterpreted as bias indicators without a clear demonstration of prejudice.
\begin{table*}[]
\tiny
\renewcommand{\arraystretch}{1.6}
\resizebox{\textwidth}{!}{%
\begin{tabular}{p{0.3\textwidth}|p{0.2\textwidth}|p{0.45\textwidth}}

News Article  & Misclassified Bias & Explanation \\
\hline
\begin{minipage}[t]{0.3\textwidth}
In Khargone district, three girls including two sisters drowned in Choral river. Most of these incidents took place when the victims were taking a dip in a river on the occasion of Sarva Pitru Amavasya. Eight persons drowned...
\end{minipage} 
& Religious Bias & \begin{minipage}[t]{0.45\textwidth}This article does not exhibit religious bias as it reports tragic drowning incidents during a religious ritual (Sarva Pitru Amavasya) without making negative judgments about the practice. The religious context is mentioned only to explain the timing of the events. It neutrally focuses on the people, locations, and circumstances involved. The purpose is to inform readers, not to promote any bias against the religion. No discrimination or bias is present in the reporting.\end{minipage}
\vspace{2pt}
\\
\hline
\begin{minipage}[t]{0.3\textwidth} 
Several railway stations and religious places in Rajasthan and Madhya Pradesh received a bomb threat after which the security around all the important places was heightened.
A letter threatening bomb...
\end{minipage} & Religious Bias & 
\begin{minipage}[t]{0.45\textwidth}
This article reports on a bomb threat in Rajasthan and Madhya Pradesh, mentioning both railway stations and religious sites without targeting any religion. The threat, attributed to Jaish-e-Mohammed, is discussed in the context of security measures and investigation. The article is neutral and focused on facts, without promoting any religious agenda.
\end{minipage} 
\vspace{2pt}
\\
\hline
\begin{minipage}[t]{0.3\textwidth} 
Former Vice President Venkaiah Naidu's office clarified that no political matters discussed during meeting with Odisha Governor Das.
 Bhubaneswar (Odisha): The office of former Vice-President M Venkaiah Naidu ...
\end{minipage} & Political Bias & 
\begin{minipage}[t]{0.45\textwidth}
This article reports on a meeting between former Vice President M. Venkaiah Naidu and Odisha Governor Raghubar Das, focusing on a clarification denying political motives. It presents a balanced view, mentioning opposition claims without endorsing or dismissing them. The tone remains neutral, avoiding political bias.
\end{minipage} 
\vspace{2pt}
\\
\hline
\begin{minipage}[t]{0.3\textwidth} 
True equality and justice require more women in politics: Rahul Gandhi

“A year ago, we launched the 'Indira Fellowship' to amplify women's voices in politics...
\end{minipage} & Political Bias & 
\begin{minipage}[t]{0.45\textwidth}
This article neutrally reports on Rahul Gandhi's call for increased women’s participation in politics and the goals of the 'Indira Fellowship' and 'Shakti Abhiyan.' It avoids endorsing or criticizing any political ideology.
\end{minipage} 
\vspace{2pt}
\\
\hline
\begin{minipage}[t]{0.3\textwidth} 
The Biju Janata Dal (BJD) on Thursday (October 3, 2024) alleged that the Odisha Raj Bhawan had turned into a ‘war room’ for the Jharkhand Assembly election. The saffron party claimed that key political manoeuvres were being orchestrated from the August address...
\end{minipage} & Regional Bias & 
\begin{minipage}[t]{0.45\textwidth}
This article does not have regional bias because it focuses on the political allegations raised by the Biju Janata Dal (BJD) regarding the involvement of Odisha Raj Bhawan in political activities related to the Jharkhand Assembly election. The article reports on the BJD’s concerns and criticisms, which are directed at specific political actions and the governor’s role, rather than making any disparaging comments or generalizations about people from a particular region.
\end{minipage}
\vspace{2pt}
\\
\hline
\begin{minipage}[t]{0.3\textwidth} 
Sun Belt officials are preparing for the possibility Texas State leaves the conference sources told the Daily News-Record on Saturday.
 Those league sources said early in the day it was ...
\end{minipage} & Regional Bias & 
\begin{minipage}[t]{0.45\textwidth}
The article analyzes potential conference realignment in college sports, focusing on logistics and geographic fit without favoring specific regions. However, mentions of historical tensions and regional rivalries could be perceived as highlighting certain areas. Overall, it provides a neutral, analytical perspective on the issue.
\end{minipage}
\end{tabular}%
}
\vspace{5pt}
\caption{LLM Misannotations Examples}
\label{tab:annotations}
\end{table*}
\end{document}